\documentclass[letterpaper]{article} 
\usepackage{aaai2026}  
\usepackage{times}  
\usepackage{helvet}  
\usepackage{courier}  
\usepackage[hyphens]{url}  
\usepackage{graphicx} 
\usepackage{natbib}  
\usepackage{caption} 
\usepackage{algorithm}
\usepackage{algorithmic}

\usepackage{amsmath}
\usepackage{pifont}
\usepackage{amsfonts}
\usepackage{multirow}
\usepackage{makecell}
\usepackage{booktabs}
\usepackage{multirow}   
\usepackage{booktabs}
\usepackage{pdfpages}
\usepackage{newfloat}
\usepackage{listings}
\DeclareCaptionStyle{ruled}{labelfont=normalfont,labelsep=colon,strut=off} 
\floatstyle{ruled}
\newfloat{listing}{tb}{lst}{}
\floatname{listing}{Listing}
\title{MonoCLUE : Object-Aware Clustering Enhances Monocular 3D Object Detection}

\author{
    Sunghun Yang, \quad
    Minhyeok Lee, \quad
    Jungho Lee, \quad
    Sangyoun Lee
}

\affiliations{
    Yonsei University\\
    Seoul, Republic of Korea\\
    sunghun98@yonsei.ac.kr, \quad hydragon516@yonsei.ac.kr, \quad 2015142131@yonsei.ac.kr, \quad  syleee@yonsei.ac.kr
}

\usepackage{bibentry}

\begin{document}

\maketitle

\begin{abstract}
Monocular 3D object detection offers a cost-effective solution for autonomous driving but suffers from ill-posed depth and limited field of view. These constraints cause a lack of geometric cues and reduced accuracy in occluded or truncated scenes. While recent approaches incorporate additional depth information to address geometric ambiguity, they overlook the visual cues crucial for robust recognition. We propose MonoCLUE, which enhances monocular 3D detection by leveraging both local clustering and generalized scene memory of visual features. First, we perform K-means clustering on visual features to capture distinct object-level appearance parts (e.g., bonnet, car roof), improving detection of partially visible objects. The clustered features are propagated across regions to capture objects with similar appearances. Second, we construct a generalized scene memory by aggregating clustered features across images, providing consistent representations that generalize across scenes. This improves object-level feature consistency, enabling stable detection across varying environments. Lastly, we integrate both local cluster features and generalized scene memory into object queries, guiding attention toward informative regions. Exploiting a unified local clustering and generalized scene memory strategy, MonoCLUE enables robust monocular 3D detection under occlusion and limited visibility, achieving state-of-the-art performance on the KITTI benchmark.

\end{abstract}

\begin{links}
\link{Code}{https://github.com/SungHunYang/MonoCLUE}
\end{links}

\section{Introduction}
3D object detection is a cornerstone of autonomous driving, enabling the estimation of the location, size, and depth of objects in a scene. To this end, LiDAR‑based, multi‑view, and monocular approaches have emerged in 3D object detection research. In particular, the simplicity of using a single image and its cost-effectiveness have made the monocular approach a topic of significant current interest. However, the lack of viewpoint disparity in monocular images leads to a loss of relative geometric cues. As a result, the model faces difficulty in projecting 2D bounding boxes into accurate 3D positions referred to as the ill-posed depth problem. Moreover, relying on a single image limits the observable field of view. Without alternative viewpoints, the model is required to infer occluded objects based solely on partial observations, which leads to reduced prediction accuracy.

\begin{figure}[t]
	\begin{center}
        \setlength{\belowcaptionskip}{-10pt}
		\includegraphics[width=\linewidth]{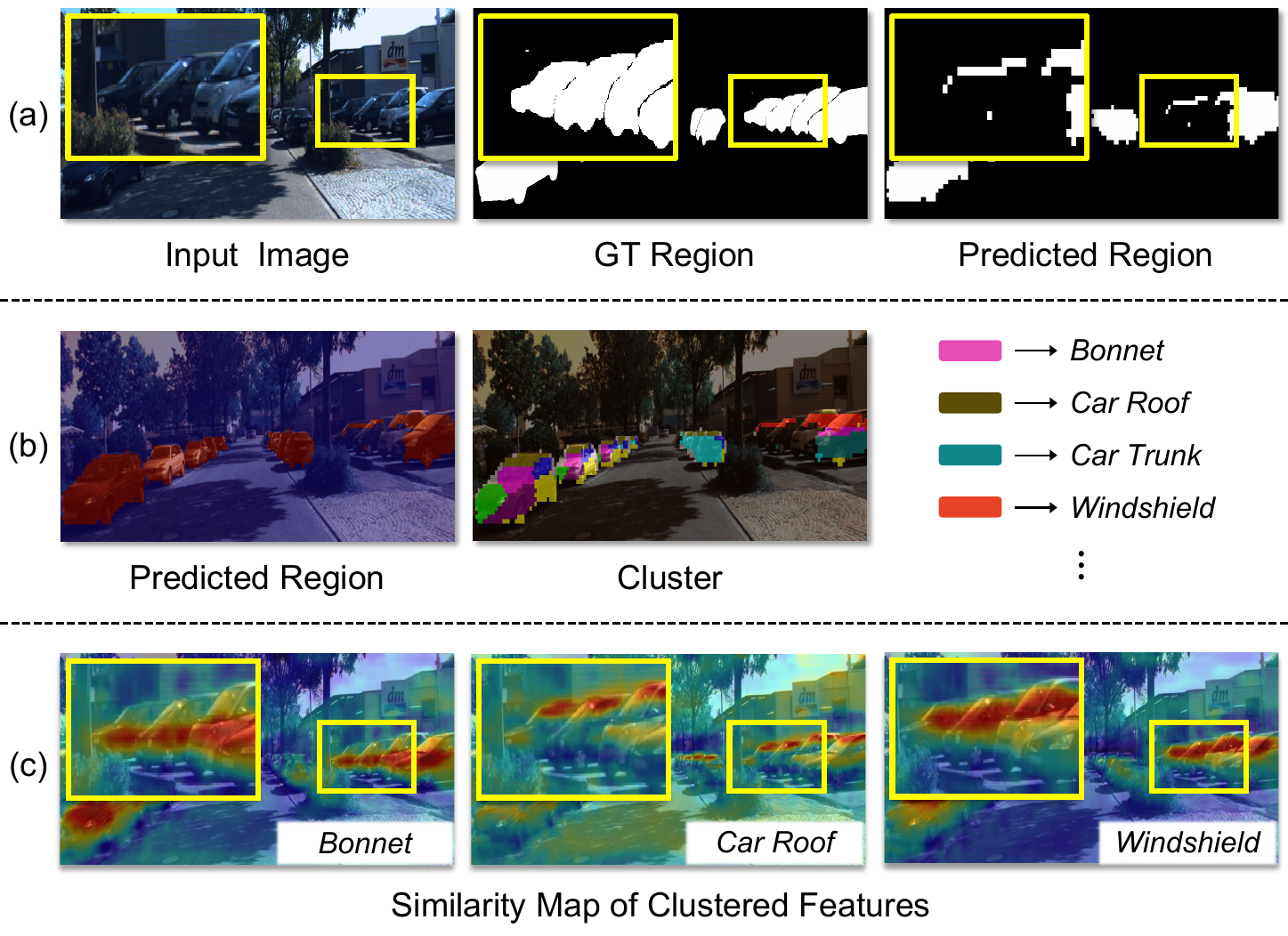}
		\caption{(a) Comparison between predicted and ground-truth regions, showing prediction errors on occluded objects. (b) Local clustering results, revealing that clusters reflect object parts and orientation, even in incomplete segmentation. (c) Activation maps of cluster features propagated across the image, capturing visually similar regions beyond initial segmentation.}
		\label{fig:fig1}
	\end{center}
\end{figure}

To address the ill-posed depth problem, previous works~\cite{zhang2023monodetr, pu2025monodgp, yan2024monocd} have explored the use of depth cues to enhance geometric reasoning. MonoDETR~\cite{zhang2023monodetr} introduces a foreground depth map to provide effective depth cues and improve geometric performance. In addition, MonoDGP~\cite{pu2025monodgp} enhances depth accuracy by incorporating geometric depth differences. These approaches enrich 3D-aware representations and mitigate geometric ambiguities using enhanced depth map.

However, these methods overlook the importance of visual cues that are fundamental to object detection. The aforementioned issue of limited observable field of view has not been the focus of previous methods. In monocular settings, key factors such as object center, spatial position, and orientation must be inferred solely from appearance. This becomes especially problematic in cases of occlusion, truncation, or complex scenes with overlapping objects. In such cases, relying on depth is insufficient to separate instances or capture their complete shape. As a result, depth-focused strategy monocular object detection remains highly challenging under these conditions. Therefore, it is crucial to emphasize diverse object-level appearance cues, which overlooked in prior works, to support reliable detection in such cases.

To solve these problems, we propose MonoCLUE.
First, we apply K-means clustering~\cite{hartigan1979algorithm} to regions of the feature map corresponding to object locations. This process aims to separate object-level appearance patterns and is referred to as local clustering. To guide this process, we utilize object segment masks generated by SAM~\cite{kirillov2023segment}, which ensure that clustering is performed within object regions. Consequently, we encourage the clustered features to represent diverse object-level visual patterns(e.g., bonnet, car roof) through training. This enables the detector to reinforce similar instances and encode object features more effectively. This method is robust to hard samples where objects are only partially visible. As shown in Figure~\ref{fig:fig1}, similar appearances are captured when the clustered features from the segmented region are propagated across the entire scene.
Second, clustering within a single image lacks the ability to capture consistent visual patterns across scenes. To complement this, we aggregate local cluster features from multiple scenes to construct a generalized scene memory that encodes commonly occurring appearance cues. This memory provides common visual patterns by reducing sensitivity to image-specific variability. Moreover, it serves as a stable reference, supporting reliable predictions when local cluster features are insufficient or ambiguous. As a result, this process enhances the overall stability of detection performance, especially in less complex scenes.
Lastly, MonoCLUE adopts the DETR~\cite{carion2020end} architecture for object detection. It further enhances the framework by integrating local cluster features  and generalized scene memory into the object queries, improving object-level reasoning. This guidance helps the queries focus on relevant regions and capture more informative object-level features. Consequently, the combined effect of local clustering and generalized scene memory leads to more robust object understanding in monocular settings.

We demonstrate through various ablation studies that our model achieves both efficiency and strong performance. This results in state-of-the-art performance in monocular settings on the KITTI benchmark~\cite{geiger2012we}.

Our main contributions are summarized as follows:

\begin{itemize}
    \item 
    We cluster local regions to capture diverse visual cues, enabling robust detection of partially visible instances and improving performance on hard samples.
    
    \item 
    We construct a generalized scene memory from aggregated local cluster features, providing consistent appearance patterns that enhance generalization across scenes.
    
    \item 
    We integrate local cluster features and generalized scene memory to the query to enhance object decoding. This leads to state-of-the-art performance on the KITTI dataset.
\end{itemize}

\section{Related Work}

\subsection{Muti-View 3D object detection}
Multi-view 3D object detection leverages images from multiple cameras to estimate object locations and shapes in 3D space. DETR3D~\cite{wang2022detr3d} introduces 3D object queries projected onto multi-view images to aggregate relevant features. BEVFormer~\cite{li2022bevformer} uses learnable BEV queries and a spatiotemporal transformer for efficient multi-view feature aggregation. Subsequent studies enhance performance with techniques like cross-modal distillation~\cite{huang2022tig}. These methods benefit from diverse viewpoints and combine geometric and appearance cues for strong performance. In contrast, monocular methods lack spatial diversity and show lower accuracy. To address this, we extract rich visual signals from a single image. MonoCLUE applies object-level clustering to enhance monocular features and strengthen object-level priors without relying on multi-view cues.

\subsection{Monocular 3D object detection}
Monocular 3D object detection estimates 3D location, dimensions, and orientation of objects from a single RGB image. Compared to multi-view–based methods, it offers a cost-efficient alternative without additional sensors. CNNs have been widely used for extracting local features and building spatial context~\cite{li2022diversity, liu2020smoke}. Many methods extend 2D detectors~\cite{wang2021fcos3d} and incorporate geometric constraints or auxiliary supervision, such as LiDAR and depth maps~\cite{ma2019accurate, reading2021categorical, wang2021depth}, to compensate for missing depth. Recently, Transformer-based models~\cite{dosovitskiy2020image} have gained attention for capturing long-range dependencies and global context. DETR-style frameworks~\cite{carion2020end} treat detection as set prediction using object queries. MonoDETR~\cite{zhang2023monodetr} introduces depth-aware queries, while MonoDGP~\cite{pu2025monodgp} improves context with segment embeddings and decoupled 2D–3D decoding.

MonoDGP uses segment embeddings to enhance context but overlooks regions outside masks and lacks feature diversity. We replace limited cues with clustering-based features to build robust representations that address monocular limitations. Our model retains query-based decoding and geometric reasoning, while enhancing appearance cues via local clustering and scene memory to improve detection.

\subsection{K-means Cluster}
K-means~\cite{hartigan1979algorithm} has been widely used in vision tasks to group features with similar patterns or semantics~\cite{guo2017deep, caron2018deep}. In object detection, it captures underlying structures within feature representations, facilitating region grouping and instance understanding. Prior works apply clustering to discover object parts~\cite{wang2020unsupervised} or for representation learning and prototype generation~\cite{caron2020unsupervised}. Inspired by this, we leverage K-means to extract diverse object-level features and enhance visual reasoning in monocular 3D detection.

\begin{figure*}[t]
	\centering
    \setlength{\belowcaptionskip}{-5pt}
	\includegraphics[width=\linewidth]{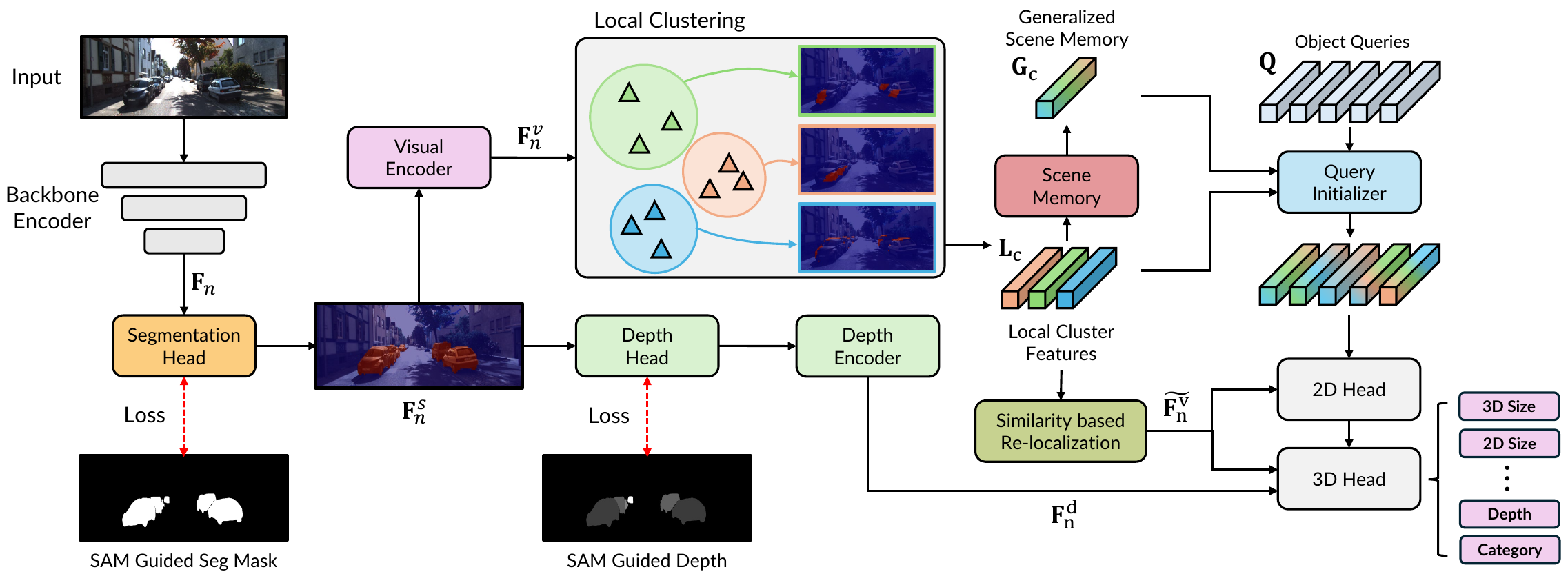}
	\caption{Overall architecture of the proposed MonoCLUE. Our core components are local clustering, similarity based re-localization, and query initialization. We perform clustering on the visual encoder features to extract local cluster features from specific regions. The local cluster features are then used for re-localization, generalized scene memory, and query initialization.}
	\label{fig:fig2}
\end{figure*}

\section{Methods}
\subsection{Overall Architecture} 
Figure~\ref{fig:fig2} presents the overall architecture of MonoCLUE. A backbone encoder extracts multi-scale features $\textbf{F}_n \in \mathbb{R}^{C \times \frac{H}{n} \times \frac{W}{n}}$, where $n \in 
\{2^3,2^4,2^5,2^6\}$. Each $\textbf{F}_n$ is processed by the region segmentation head. Its output is supervised using SAM~\cite{kirillov2023segment}-guided segmentation masks. The seg-embedded features $\textbf{F}_n^s$ produced by the region segmentation head are subsequently passed through both the visual and the depth encoder, following the structure of MonoDETR~\cite{zhang2023monodetr}.

Firstly, we obtain visual features $\mathbf{F}_{n}^{v}$, which are the output of the visual encoder. Afterward, these features are clustered by K-means clustering within object-region segmentation mask as shown in Figure~\ref{fig:fig2}. We define each clustered region, which captures diverse visual cues, as local cluster features $\textbf{L}_c \in \mathbb{R}^{N_l \times C}$, where $N_l$ is the number of clusters. To enhance region identification, we additionally perform similarity-based re-localization. This involves identifying regions that show high similarity with the $\textbf{L}_c$. Such a process is utilized to assist the model in discovering object-like areas, including partially visible objects. Secondly, we gather all $\textbf{L}_c$ and store them in a generalized scene memory $\textbf{G}_c \in \mathbb{R}^{N_g \times C}$, where $N_g$ represents the number of generalized scene memories. These memories, which are learned dataset-wide representations, capture common appearance patterns. Lastly, we use object queries $\textbf{Q} \in \mathbb{R}^{N_q \times C}$, where $N_q$ denotes the number of queries, to decode the learned information. $\textbf{Q}$ is directly initialized with both $\textbf{L}_c$ and $\textbf{G}_c$ before decoding. This initialization allows $\textbf{Q}$ to embed object-aware features in advance. Then, following the structure of MonoDGP~\cite{pu2025monodgp}, we decode initialized $\textbf{Q}$ with separate 2D and 3D heads to perform 3D object detection. The detailed descriptions of each component are provided in the following sections.

\subsection{Local Clustering} 
Figure~\ref{fig:fig3}(a) shows the process of our local clustering. The visual encoder feature $\textbf{F}_n^v$ contains information required for both object classification and box regression. Especially in 3D object detection, since visual characteristics change depending on the orientation and depth of an object, $\textbf{F}_n^v$ inevitably includes  2D and 3D-aware information. To make these representations more distinguishable, we cluster $\textbf{F}_n^v$ to explicitly separate and emphasize distinct visual cues. 

To this end, we replace the box-shaped masks used in MonoDGP with object-shaped masks to supervise the segmentation head. This allows clustering to focus exclusively on object regions. As a result, the quality of the clusters is significantly enhanced by eliminating background noise, leading to improved discrimination. Based on this strategy, we apply K-means clustering to $\textbf{F}_n^v$ exclusively within the segmentation mask $M_n \in \mathbb{R}^{\frac{H}{n} \times \frac{W}{n}}$ predicted from the region segmentation head. We then apply masked average pooling to each of the $N_l$ clusters to obtain the local cluster features $\textbf{L}_c$. This process is expressed as follows:
\begin{equation}
\begin{aligned}
    \textbf{L}_c^{(k)} = \frac{\sum_{i,j} M_n^{(k)}(i,j) \cdot \textbf{F}_n^v(i,j)}{\sum_{i,j} M_n^{(k)}(i,j)} \quad \text{for } k = 1, \dots, N_l,
\end{aligned}
\end{equation}
where $(i, j)$ are the pixel coordinates. Consequently, $\textbf{L}_c$ exhibits enhanced reliability in specific regions of partially visible objects, as the separation of features enables accurate discrimination even under partial visibility conditions

\subsection{Generalized Scene Memory}
\begin{figure}[t]
	\begin{center}
        \setlength{\belowcaptionskip}{-15pt}
		\includegraphics[width=\linewidth]{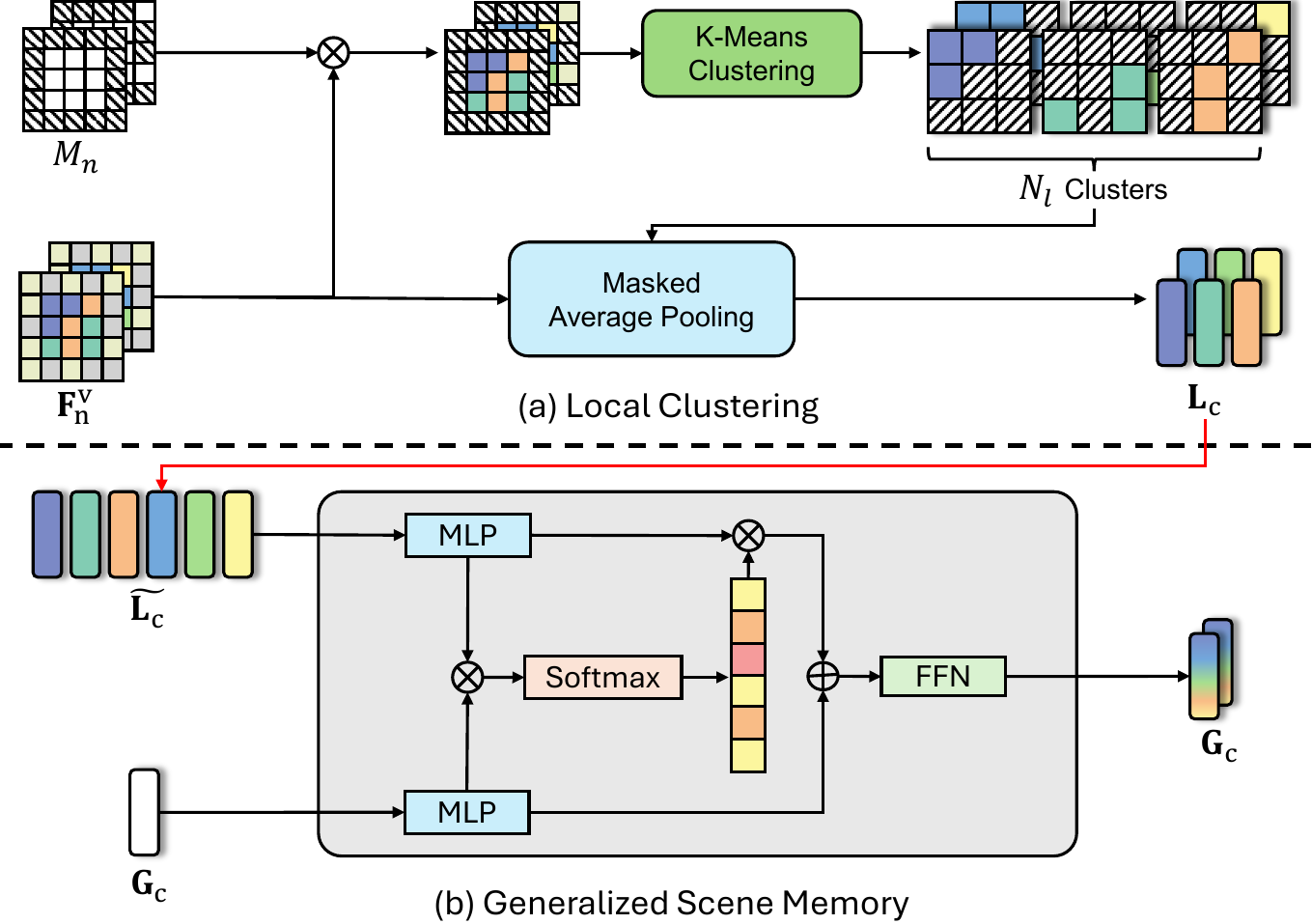}
		\caption{The process of local clustering and generalized scene memory. Each feature corresponds to a monocular image. (a) Local cluster features are obtained by independently extracting $N_l$ features from the masked object regions. (b) Generalized scene memory integrates $\textbf{L}_c$ from multiple images into a shared representation.}
		\label{fig:fig3}
	\end{center}
\end{figure}

Since $\textbf{L}_c$ comes from single-image clustering, it may lack generalization capability. To complement such image-specific features, it is necessary to capture general object priors. Therefore, we introduce a generalized scene memory. Figure~\ref{fig:fig3}(b) illustrates the structure of the generalized scene memory procedure. The generalized scene memories $\textbf{G}_c$ store common and recurring object features across the dataset. By collecting image-specific $\textbf{L}_c$ from all images and extracting shared features, we obtain a generalized representation for the dataset. 

To this end, we first create $N_g$ embedding vectors as memory. We then incorporate $\textbf{L}_c$ into them using a cross-attention mechanism~\cite{vaswani2017attention}, where the memory vectors $\textbf{G}_c$ serve as query. We apply this process at every training iteration to progressively update the memory, encouraging $\textbf{G}_c$ to store useful features that are commonly shared across the dataset. Cross-attention enables $\textbf{G}_c$ to aggregate diverse $\textbf{L}_c$ into general object-level cues in a balanced manner. In training stage, to ensure consistency across batches, the same memories $\textbf{G}_c$ are shared for all inputs. Therefore, we flatten the batch dimension of $\textbf{L}_c$, resulting in $ \tilde{\mathbf{L}}_{c} \in  \mathbb{R}^{(B \times N_l) \times C}$, and use it as the key and value in the cross-attention. This process is expressed as follows:
\begin{equation}
\begin{aligned}
    \textbf{G}_c = \textrm{softmax}\left(\frac{w_q \textbf{G}_c (w_k \tilde{\mathbf{L}}_c)^\top}{\sqrt{C}} \right) (w_v \tilde{\mathbf{L}}_c) + w_q\textbf{G}_c,
\end{aligned}
\end{equation}
where $w_q$,$w_k$ and $w_v$ are learnable projection matrices of query, key and value.
By integrating a dataset-wide shared features into $\textbf{G}_c$, the model obtains a generalized object representation that reflects common appearance patterns. These generalized priors remain effective even in unseen scenes and contribute to stabilizing predictions. In addition, $\textbf{G}_c$ perform robustly on easy objects that resemble frequently seen prototypes during training.

\subsection{Similarity-based Re-localization}
In general, the segmentation head tends to produce inaccurate masks in cases of occlusion or small objects due to insufficient visual cues. Such errors degrade the final 3D object detection performance. 

\begin{figure}[t]
	\begin{center}
        \setlength{\belowcaptionskip}{-8pt}
		\includegraphics[width=\linewidth]{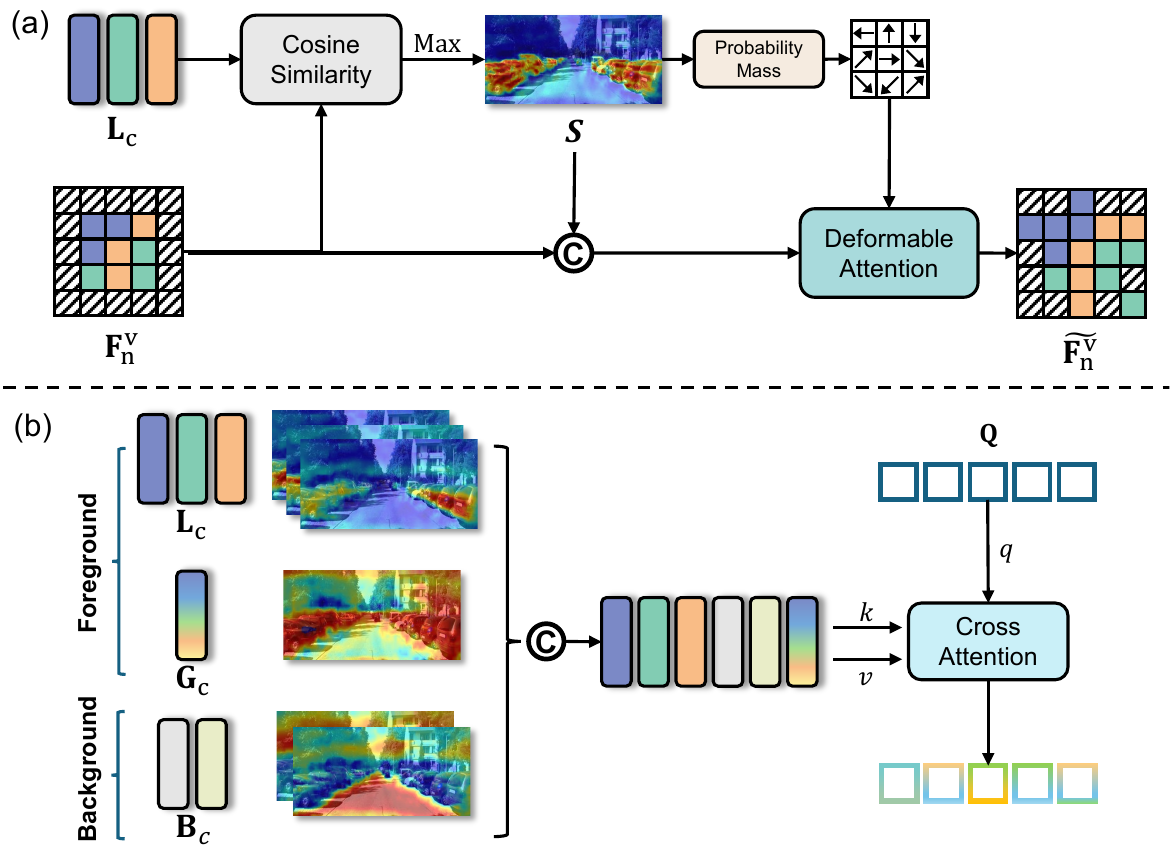}
		\caption{The process of re-localization and query initialization. (a) Local cluster similarity is used to compute $\textbf{S}$, which re-localizes features to object-like regions. (b) Query initialization using $\textbf{L}_c$, $\textbf{G}_c$, and $\textbf{B}_c$.}
		\label{fig:fig4}
	\end{center}
\end{figure}

To address this, we utilize the $\textbf{L}_c$ to re-localize and refine object regions based on high similarity.
This approach enables broader discovery of objects that exhibit sparse visual cues. First, as shown in Figure~\ref{fig:fig4}(a), we calculate pixel-wise cosine similarity scores for $\textbf{F}_n^v$ against all $N_l$ instances of $\textbf{L}_c$. Therefore, the resulting $N_l$ local similarity maps has dimensions of $N_l\times\frac{H}{n} \times \frac{W}{n}$. Next, we take the maximum value along the $N_l$ dimension of this local similarity maps to generate final similarity map $\textbf{S} \in \mathbb{R}^{\frac{H}{n} \times \frac{W}{n}}$. This final $\textbf{S}$ identifies object-like candidate regions based on high similarity to any of the $N_l$ clustered features. This process is expressed as follows: 
\begin{equation}
\begin{aligned}
    S(i, j) = \max_{N_l} \left( \frac{ \mathbf{L}_c \cdot \mathbf{F}_n^v(i, j) }{ \left\| \mathbf{L}_c \right\| \left\| \mathbf{F}_n^v(i, j) \right\| } \right).
\end{aligned}
\end{equation}
The $\textbf{S}$ helps re-localize the candidate positions of objects that are not detected due to insufficient visual cues. Second, the $\textbf{S}$ is concatenated with $\textbf{F}_n^v$ to inject candidate object location cues into the features, guiding the model to focus on likely object regions. In this method, we employ the Multiscale Deformable attention (MsDeform)~\cite{zhu2020deformable} for memory efficiency, similar to existing methods. As a result, we generate refined features $\tilde{\textbf{F}}_n^v$, which enhances the model's localization performance. 
Lastly, we guide MS-Deform attention to focus on object-centric regions, which helps prevent attention weights from shifting toward the background due to randomly initialized offsets. To achieve this, we initialize reference point offsets using the similarity map $\textbf{S}$, which serves as an auxiliary guide. In this process, the softmax correlation map is interpreted as a probability mass over the reference lattice $r$~\cite{luvizon2019human}. A weighted average of grid locations estimates the object center $c$, and subtracting $r$ from $c$. This yields coarse offsets that bias sampling toward object-centric regions. This process is expressed as follows:
\begin{gather}
c = \sum_{i,j} \textrm{softmax}(\textbf{S}(i,j)) \cdot r(i,j), \\
\Delta(i,j) = c - r(i,j).
\end{gather}
The detailed MS-Deform attention weight results are provided in the supplementary material.

\subsection{Query Initializer}
Following baseline~\cite{pu2025monodgp}, we use object queries $\textbf{Q}$ for decoding and process them through a 2D and a 3D head. These queries receive information from the refined visual feature $\tilde{\textbf{F}_n^v}$ and the depth encoder output $\textbf{F}_n^d$ via cross-attention, which assigns higher weights to more similar features during aggregation.

To guide this process more effectively, we pre-inject the diverse visual cues from $\textbf{L}_c$ and $\textbf{G}_c$ into the $\textbf{Q}$. This initialization offers object-aware priors, allowing $\textbf{Q}$ to better capture visual representations relevant to the target object.
Since decoding requires not only object-level understanding but also contextual awareness, background features become essential.
These features provide complementary cues that help refine object boundaries and improve 3D localization.
To leverage this, we cluster the background excluding the object masks ($1 - M_n$) into $N_b$ groups to obtain background feature $\textbf{B}_c \in \mathbb{R}^{N_b \times C}$ using the same method as for the local clustering. As noted in~\cite{yang2024monogae, yang2023gedepth}, depth is closely related to ground surface, incorporating these cues contributes to accurate 3d object decoding. Lastly, as shown in Figure~\ref{fig:fig4}(b), we incorporate $\textbf{B}_c$ alongside $\textbf{L}_c$ and $\textbf{G}_c$ during query initialization. Since these features already summarize meaningful object and scene-level information, there is no need to attend to the entire spatial feature map. Instead of applying cross-attention directly to the full feature map with spatial dimensions of $\frac{H}{n} \times \frac{W}{n}$ per level, we use a compact set of representative features. Specifically, we select $N_l + N_g + N_b$ features that effectively summarize the representation. This reduces memory consumption and enables more efficient attention computation.

As a result, embedding both object-aware and context-aware cluster information into $\textbf{Q}$ leads to more robust and accurate object predictions.

\subsection{Loss function}
We follow the loss formulation of MonoDGP, which includes region segmentation loss $\mathcal{L}_{\text{region}}$, depth map regression loss $\mathcal{L}_{\text{depth}}$, 2D detection loss $\mathcal{L}_{\text{2D}}$, and 3D object estimation loss $\mathcal{L}_{\text{3D}}$. Here, $\mathcal{L}_{2D}$ covers classification, 2D bounding box regression, GIoU, and projected center losses. While, $\mathcal{L}_{3D}$ handles 3D size, orientation, and center depth. The total loss is expressed as follows:
\begin{gather}
    \mathcal{L}_{\text{total}} = \mathcal{L}_{\text{2D}} + \mathcal{L}_{\text{3D}} + \lambda \mathcal{L}_{\text{depth}} + \lambda \sum_{i=0}^{4} \mathcal{L}_{\text{region}}^i,
\end{gather}
where $\lambda$ controls the depth and region segmentation loss weights.

\begin{table*}[!ht]
\small
\setlength{\tabcolsep}{3pt}
\centering
\renewcommand{\arraystretch}{0.9}
\setlength{\belowcaptionskip}{-12pt}
\begin{tabular}{l||c||ccc|ccc||ccc|ccc}
\toprule
\multirow{3}{*}{\textbf{Methods}} & \multirow{3}{*}{\textbf{Extra}} 
& \multicolumn{6}{c||}{\textbf{Test}} 
& \multicolumn{6}{c}{\textbf{Validation}} \\
\cmidrule{3-14}
&& \multicolumn{3}{c|}{$AP_{BEV|R40}$} & \multicolumn{3}{c||}{$AP_{3D|R40}$}
& \multicolumn{3}{c|}{$AP_{BEV|R40}$} & \multicolumn{3}{c}{$AP_{3D|R40}$} \\
&& Easy & Mod. & Hard & Easy & Mod. & Hard & Easy & Mod. & Hard & Easy & Mod. & Hard \\
\midrule\midrule

MonoPGC~\cite{wu2023monopgc} & \multirow{2}{*}{Depth} & 32.50 & 23.14 & 20.30 & 24.68 & 17.17 & 14.14 & 34.06 & 24.26 & 20.78 & 25.67 & 18.63 & 15.65 \\
OPA-3D~\cite{su2023opa} & & 33.54 & 22.53 & 19.22 & 24.60 & 17.05 & 14.25 & 33.80 & 25.51 & 22.13 & 24.97 & 19.40 & 16.59 \\
\addlinespace[0.5pt]
\midrule

MonoDTR~\cite{huang2022monodtr} & \multirow{3}{*}{LiDAR} & 28.59 & 20.38 & 17.14 & 21.99 & 15.39 & 12.73 & 33.33 & 25.35 & 21.68 & 24.52 & 18.57 & 15.51 \\
DID-M3D~\cite{peng2022did} & & 32.95 & 22.76 & 19.83 & 24.40 & 16.29 & 13.75 & 31.10 & 22.76 & 19.50 & 22.98 & 16.12 & 14.03 \\
OccupancyM3D~\cite{peng2024learning} & & 35.38 & 24.18 & 21.37 & 25.55 & 17.02 & 14.79 & 35.72 & 26.60 & 23.68 & 26.87 & 19.96 & 17.15 \\
\addlinespace[0.5pt]
\midrule
GUPNet~\cite{lu2021geometry} & \multirow{10}{*}{-} & 30.29 & 21.19 & 18.20 & 22.26 & 15.02 & 13.12 & 31.07 & 22.94 & 19.75 & 22.76 & 16.46 & 13.72 \\
MonoCon~\cite{liu2022learning} &  & 31.12 & 22.10 & 19.00 & 22.50 & 16.46 & 13.95 & - & - & - & 26.33 & 19.01 & 15.98 \\
DEVIANT~\cite{kumar2022deviant} & & 29.65 & 20.44 & 17.43 & 21.88 & 14.46 & 11.89 & 32.60 & 23.04 & 19.99 & 24.63 & 16.54 & 14.52 \\
MonoDDE~\cite{li2022diversity} & & 33.58 & 23.46 & 20.37 & 24.93 & 17.14 & 15.10 & 35.51 & 26.48 & 23.07 & 26.66 & 19.75 & 16.72 \\
MonoUNI~\cite{jinrang2023monouni}& & 33.28 & 23.05 & 19.39 & 24.75 & 16.73 & 13.49 & - & - & - & 24.51 & 17.18 & 14.01 \\

MonoDETR~\cite{zhang2023monodetr} & & 33.60 & 22.11 & 18.60 & 25.00 & 16.47 & 13.58 & 37.86 & 26.95 & 22.80 & 28.84 & 20.61 & 16.38 \\
FD3D~\cite{wu2024fd3d} & & 34.20 & 23.72 & 20.76 & 25.38 & 17.12 & 14.50 & 36.98 & 26.77 & 23.16 & 28.22 & 20.23 & 17.04 \\
MonoMAE~\cite{jiang2024monomae} & & 34.15 & 24.93 & 21.76 & 25.60 & \underline{18.84} & \textbf{16.78} & \underline{40.26} & 27.08 & 23.14 & 30.29 & 20.90 & 17.61 \\
MonoCD~\cite{yan2024monocd} & & 33.41 & 22.81 & 19.57 & 25.53 & 16.59 & 14.53 & 34.60 & 24.96 & 21.51 & 26.45 & 19.37 & 16.38 \\

MonoDGP~\cite{pu2025monodgp} & & \underline{35.24} & \underline{25.23} & \underline{22.02} & \underline{26.35} & 18.72 & 15.97 & 39.40 & \underline{28.20} & \underline{24.42} & \underline{30.76} & \underline{22.34} & \underline{19.02} \\
\addlinespace[0.5pt]
\midrule\midrule

MonoCLUE & - & \textbf{36.15} & \textbf{26.15} & \textbf{22.81} & \textbf{27.94} & \textbf{19.70} & \underline{16.69} & \textbf{41.79} & \textbf{29.91} & \textbf{26.00} & \textbf{33.74} & \textbf{24.10} & \textbf{20.58} \\
\bottomrule
\end{tabular}

\caption{Comparisons with monocular methods on the KITTI validation and test for the car category. The best results are shown in bold, and the second-best are shown with underline.}
\label{tab:tab1}
\end{table*}

\section{Experiments}

\subsection{Dataset}

\subsubsection{KITTI}
We conduct experiments on the widely used KITTI benchmark~\cite{geiger2012we}, which is a standard dataset for 3D object detection in autonomous driving. It consists of 7,481 training and 7,518 test images, with annotations for three object categories, namely Car, Pedestrian, and Cyclist. Each object instance is assigned one of three difficulty levels Easy, Moderate and Hard based on factors such as occlusion and truncation. Following common practice~\cite{chen20153d}, we split the training set into 3,712 training and 3,769 validation images for ablation and comparison. We report performance using Average Precision $AP$ metrics for both 3D bounding boxes $AP_{3D}$ and bird's eye view $AP_{BEV}$, evaluated at 40 recall positions for each difficulty level.

\subsection{Implementation Details}
We adopt ResNet-50~\cite{he2016deep} as the backbone to extract multi-scale features. Following DETR-based designs, we use 50 object queries represented as learnable embeddings, and each attention module employs 8 heads. For memory efficiency, we apply multi-scale deformable attention~\cite{zhu2020deformable} in both the visual encoder and re-localization, using 4 sampling points. For depth prediction, the object-shaped depth range (0–60m) is uniformly quantized into 80 bins. To accelerate clustering, we use a CUDA-implemented K-means~\cite{geiger2012we} algorithm for efficient GPU computation. We set $N_l$, $N_g$, and $N_b$ to 10, the number of classes, and 3, respectively. Our model is trained on a single RTX 3090 GPU for 250 epochs with batch size 8, using AdamW~\cite{loshchilov2017decoupled} with initial learning rate $2 \times 10^{-4}$ and step decay schedule. During inference, queries with confidence below 0.2 are filtered, and no post-processing like NMS is applied.

\subsection{Results on KITTI}
We evaluate MonoCLUE on the KITTI benchmark and compare it with recent monocular 3D detection methods. As shown in Table~\ref{tab:tab1}, MonoCLUE achieves the best performance across all difficulty levels on both the test and validation sets for the car category, except for the hard case on the test set. This is achieved without using any extra information like depth or LiDAR. On the test set, it achieves 27.94 and 19.70 $AP_{3D}$ for the easy and moderate cases, outperforming previous state-of-the-art methods by +1.59\% and +0.86\%, respectively. On the validation set, it further improves performance by +2.98\% (easy) and +1.76\% (moderate). These gains are attributed to our structured clustering approach and generalized memory, which together enhance object-level discrimination and improve localization accuracy. Table~\ref{tab:tab1} highlights the effectiveness of MonoCLUE in capturing discriminative visual cues under monocular settings. Additional experiments on the KITTI dataset are included in the supplementary material.

\subsection{Results on Other Categories}
Table~\ref{tab:tab2} shows that MonoCLUE achieves the best performance for Pedestrian and the second-best for Cyclist on the KITTI test set. This indicates the effectiveness of our method across other object types. Furthermore, both the detection and orientation metrics show substantial improvements over previous methods. These results show that our enhanced visual cues contribute to improved 2D detection performance. Moreover, when geometric reasoning is incorporated in the 3D detection, these results further boost overall performance. In addition, since orientation in monocular settings relies heavily on visual cues, our diverse visual patterns prove effective in ensuring robustness to orientation.

\begin{table}[]
\small
\setlength{\tabcolsep}{2pt}
\renewcommand{\arraystretch}{0.9}
\setlength{\belowcaptionskip}{-10pt}
\begin{tabular}{l||cccc}
\toprule
Method   & Detection & Orientation & \begin{tabular}[c]{@{}c@{}}Pedestrain\\ $AP_{3D|Mod.}$\end{tabular} & \begin{tabular}[c]{@{}c@{}}Cyclist\\ $AP_{3D|Mod.}$\end{tabular} \\
\cmidrule(lr){1-5}
GUPNet  & 94.15     & 93.92       & 9.76                                                       & \textbf{3.21}                                                   \\
DEVIANT  & \underline{94.42}     & 94.01       & 8.65       & 3.13    \\
MonoDGP  & 94.35     & \underline{94.22}       & \underline{9.89}   & 2.28    \\ \midrule
MonoCLUE & \textbf{95.82}    & \textbf{95.54}      & \textbf{10.45}   & \underline{3.20}\\
\bottomrule
\end{tabular}
\caption{Comparison of multi-category, 2D detection, and orientation on the test set. The detection and orientation results are evaluated on the car category.}
\label{tab:tab2}
\end{table}

\subsection{Qualitative results}
We compare the $3D$ and $BEV$ detection performance of our method with two baseline models, MonoDETR and MonoDGP, on the KITTI validation set.
As shown in Figure~\ref{fig:fig5}, MonoCLUE exhibits more reliable overall detection quality than the baselines. Specifically, our method demonstrates more robust detection on small and distant objects that are occluded. This improvement is attributed to the cluster features, which re-localize objects by matching them to easily detectable patterns across the image. This allows the model to infer missing object parts based on similar appearances observed in less challenging regions. Additionally, in the last example, a row of aligned cars shares similar orientation. This leads to similar appearances, which result in similar cluster features. Therefore, these similar cluster features lead to consistent detection in the $BEV$. As a result, MonoCLUE demonstrates robust performance in challenging scenarios such as occlusion, truncation, and overlapping objects. Additional activation map and qualitative results are provided in the supplementary material.

\begin{table}[]
\small
\renewcommand{\arraystretch}{1.0}
\setlength{\tabcolsep}{2pt}
\setlength{\belowcaptionskip}{-5pt}
\begin{tabular}{c||ccc}
\toprule
\multirow{2}{*}{Architecture} & \multicolumn{3}{c}{$AP_{3D|R40}$} \\
\cmidrule(lr){2-4}
                              & \raisebox{-0.5ex}{Easy}   & \raisebox{-0.5ex}{Moderate}   & \raisebox{-0.5ex}{Hard}   \\ \midrule
None                          &30.66       &23.03           &19.71       \\
Codebook                      &31.77 ( \underline{+1.11} )   &23.22 ( \underline{+0.19} )      &19.75 ( \underline{+0.04} )       \\
Cross attention               &33.74 ( \underline{+3.08} )   &24.10 ( \underline{+1.07} )      &20.58 ( \underline{+0.87} )     \\ \bottomrule

\end{tabular}
\caption{Comparison of generalized scene memory architecture on the validation set. Underlined score indicates the comparison without using the generalized scene memory.}
\label{tab:tab3}
\end{table}

\section{Ablation Analysis}

\begin{table}[]
\centering
\small
\setlength{\tabcolsep}{2pt}
\renewcommand{\arraystretch}{1.0}
\setlength{\belowcaptionskip}{-12pt}
\begin{tabular}{ccc||ccc}
\toprule
\multirow{2}{*}{\shortstack{SAM\\Guidance}} & \multirow{2}{*}{\shortstack{Query\\Initializer}} & \multirow{2}{*}{\shortstack{Similarity-based\\Re-localization}} & \multicolumn{3}{c}{AP$_{3D|R40}$} \\
\cmidrule(lr){4-6}
 & & &Easy &Moderate &Hard \\
\midrule
- & - & - & 29.61 & 22.06 &18.75 \\
\checkmark & - & - & 29.82 & 22.62 & 19.30 \\
\checkmark & \checkmark & - & \underline{32.91} & \underline{23.93} & \underline{20.36} \\
\checkmark & - & \checkmark & 31.14 & 23.20 & 20.02 \\
\checkmark & \checkmark & \checkmark & \textbf{33.74} & \textbf{24.10} & \textbf{20.58} \\
\bottomrule
\end{tabular}
\caption{Comparison of core components on the validation set. Each result includes the checked component along with all other fixed components.}
\label{tab:tab4}
\end{table}

\begin{table}[]
\centering
\setlength{\tabcolsep}{2pt}
\setlength{\belowcaptionskip}{-15pt}
\renewcommand{\arraystretch}{1.0}
\begin{tabular}{l||cccc}
\toprule
Method  & \begin{tabular}[c]{@{}c@{}}Params\\ (M)\end{tabular} & \begin{tabular}[c]{@{}c@{}}FLOPs $\downarrow$ \\ (G)\end{tabular} & \begin{tabular}[c]{@{}c@{}}AP$_{3D|R40}$ $\uparrow$ \\ (Moderate)\end{tabular} &\begin{tabular}[c]{@{}c@{}}Runtime $\downarrow$ \\(ms)\end{tabular} \\
\midrule
MonoDETR & 37.68        & 59.72       & 20.61    & 35  \\ 
MonoDGP  & 42.16           & 68.99     & 22.34  & 42  \\
MonoCLUE & 44.17      & 72.71   & 24.10  & 52  \\ 

\bottomrule
\end{tabular}
\caption{Comparison of computational complexity. FLOPs and Runtime are measured on a single RTX 3090 GPU with a batch size of 1.}
\label{tab:tab5}
\end{table}

\begin{figure*}[t]
	\centering
    \setlength{\belowcaptionskip}{-8pt}
	\includegraphics[width=\linewidth]{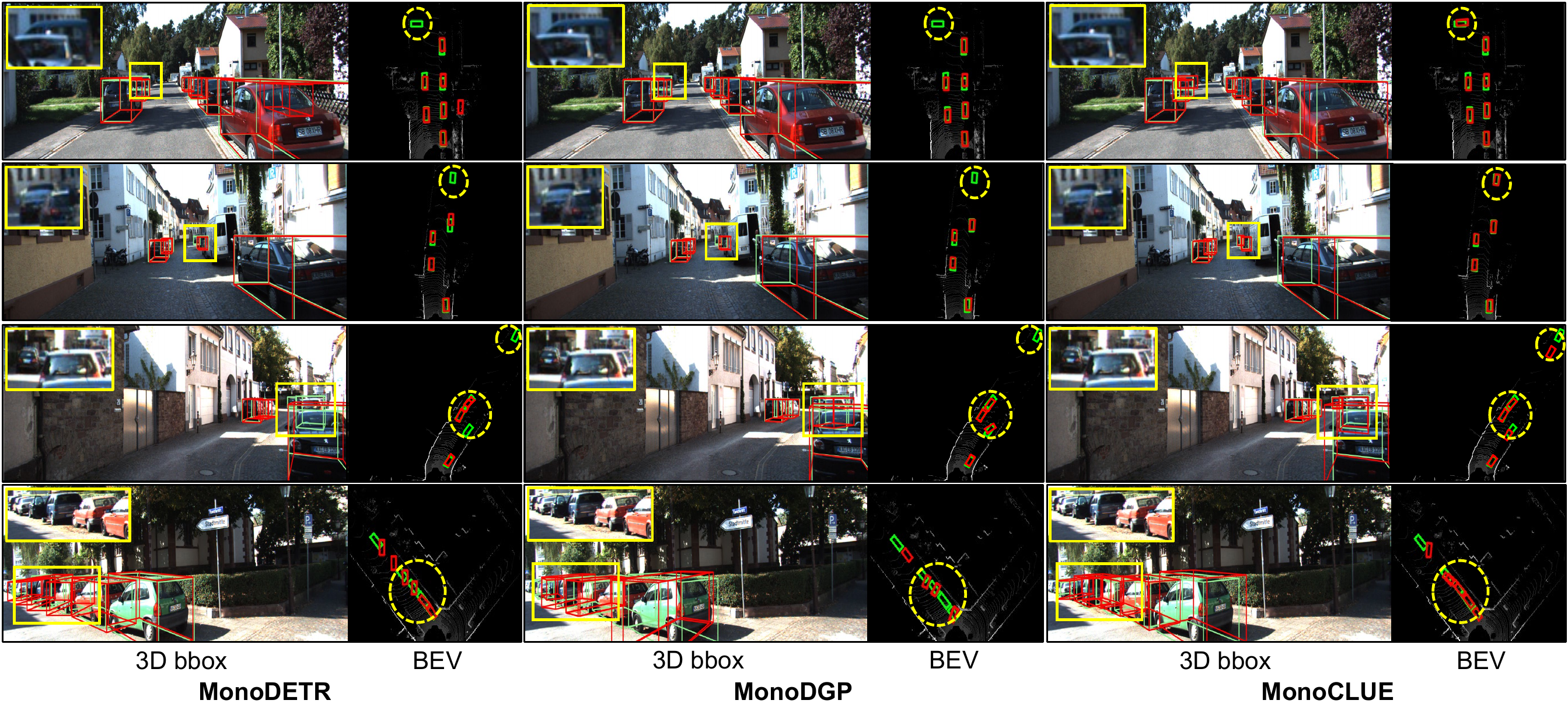}
	\caption{Qualitative comparison on the KITTI validation set. Ground-truth boxes (green) and predictions (red) are shown for both 3D bounding boxes and bird’s-eye view (BEV).}
	\label{fig:fig5}
\end{figure*}

\subsection{Scene memory}
Table~\ref{tab:tab3} demonstrates the effectiveness of the proposed generalized scene memory. The results show that performance is lowest when the generalized scene memory is not used, and the results align with our intention, as the easy case shows the most significant improvement. Specifically, the gains are +3.08\% and +1.11\%, confirming that the cross-attention-based memory is the most effective.
To explore optimal memory designs, we employ the codebook proposed in VQ-VAE~\cite{van2017neural}, a commonly used memory structure. All memory types are configured with the same size for fair comparison. However, the codebook structure struggle to find representative features using only the loss as guidance. Additionally, since not all codebook vectors are updated during training, some memory slots remained unused, leading to lower performance. In contrast, the cross-attention structure applies weights to all memory entries and learns an common feature, resulting in the best performance.

\subsection{Core components}
Table~\ref{tab:tab4} analyzes the contribution of each component in MonoCLUE. Removing SAM guidance corresponds to training with box-shaped masks, as in baseline methods~\cite{pu2025monodgp}. This hinders clustering by including background noise in object regions, which degrades performance. This highlights the importance of SAM for effective clustering. Re-localization improves occluded region representation by identifying candidate locations with similar appearances through local clustering. This leads to a performance gain of +0.7\% in the hard case. The query initializer aggregates all clustering information, leading to consistent improvements across difficulty levels. Compared to using only SAM, it achieves gains of +3.9\% in easy and +1.31\% in moderate case, validating the effectiveness of combining background and generalized information. Finally, integrating all components yields the highest performance, demonstrating the complementary benefits of each component.

\subsection{Efficiency and Performance Comparison}
Table~\ref{tab:tab5} presents a comparison of each method in terms of model complexity, computational complexity and moderate level performance $AP_{3D}$ on the KITTI validation set. Compared to the baseline MonoDETR, MonoDGP improves the performance by +1.73\% compared to our baseline, but at the cost of a significant increase in parameters 4.48M and FLOPs 9.27G. In contrast, our method MonoCLUE achieves a larger performance gain of +1.76\%, while incurring a much smaller increase in parameters 2.01M and FLOPs 3.72G. This indicates that the clustering-based design enables performance improvements without incurring significant computational or parameter overhead. As a result, MonoCLUE achieves a better cost-performance trade-off than previous state-of-the-art models.

\section{Conclusion}
We propose MonoCLUE, a monocular 3D object detection framework that improves object-level reasoning through localized clustering and scene priors. By clustering fine-grained features in object-shaped regions and generalizing them into a memory module, our method captures divers object patterns. Incorporating local, background, and scene-level features enables informative query initialization with strong contextual and geometric cues. Experiments on the KITTI benchmark validate the effectiveness of each component, with MonoCLUE achieving state-of-the-art performance among monocular methods.

\section*{Acknowledgements}
This work was supported by the Korea Institute of Science and Technology (KIST) Institutional Program (Project No.2E33612-25-016) and by the National Research Foundation of Korea (NRF) grant funded by the Korea government (MSIT)(No. RS-2024-00340745). 

\bibliography{aaai2026}

\clearpage
\includepdf[pages=-]{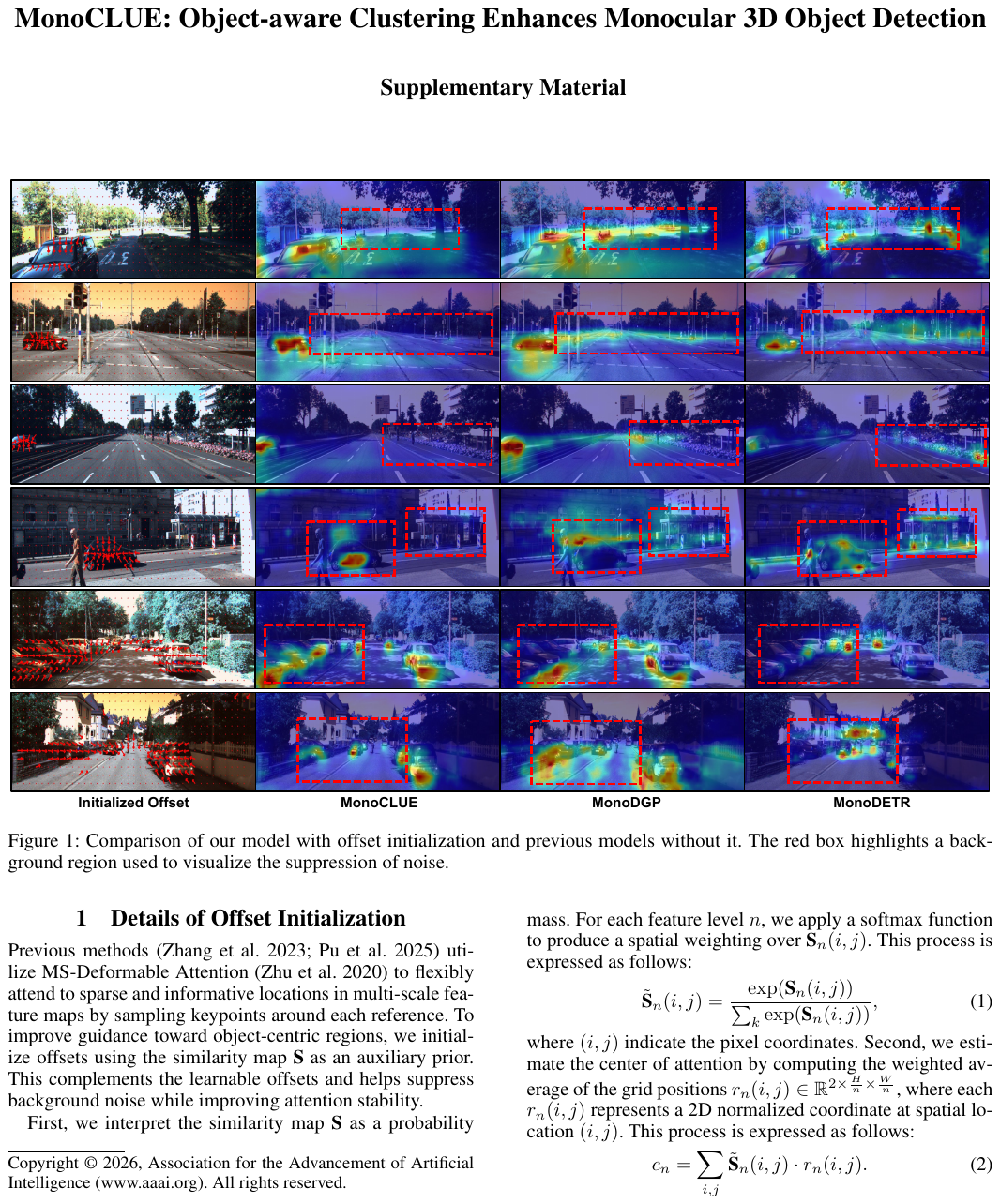}

\end{document}


\maketitle

\begin{strip}
    \vspace{-70pt}
	\centering
	\includegraphics[width=\linewidth]{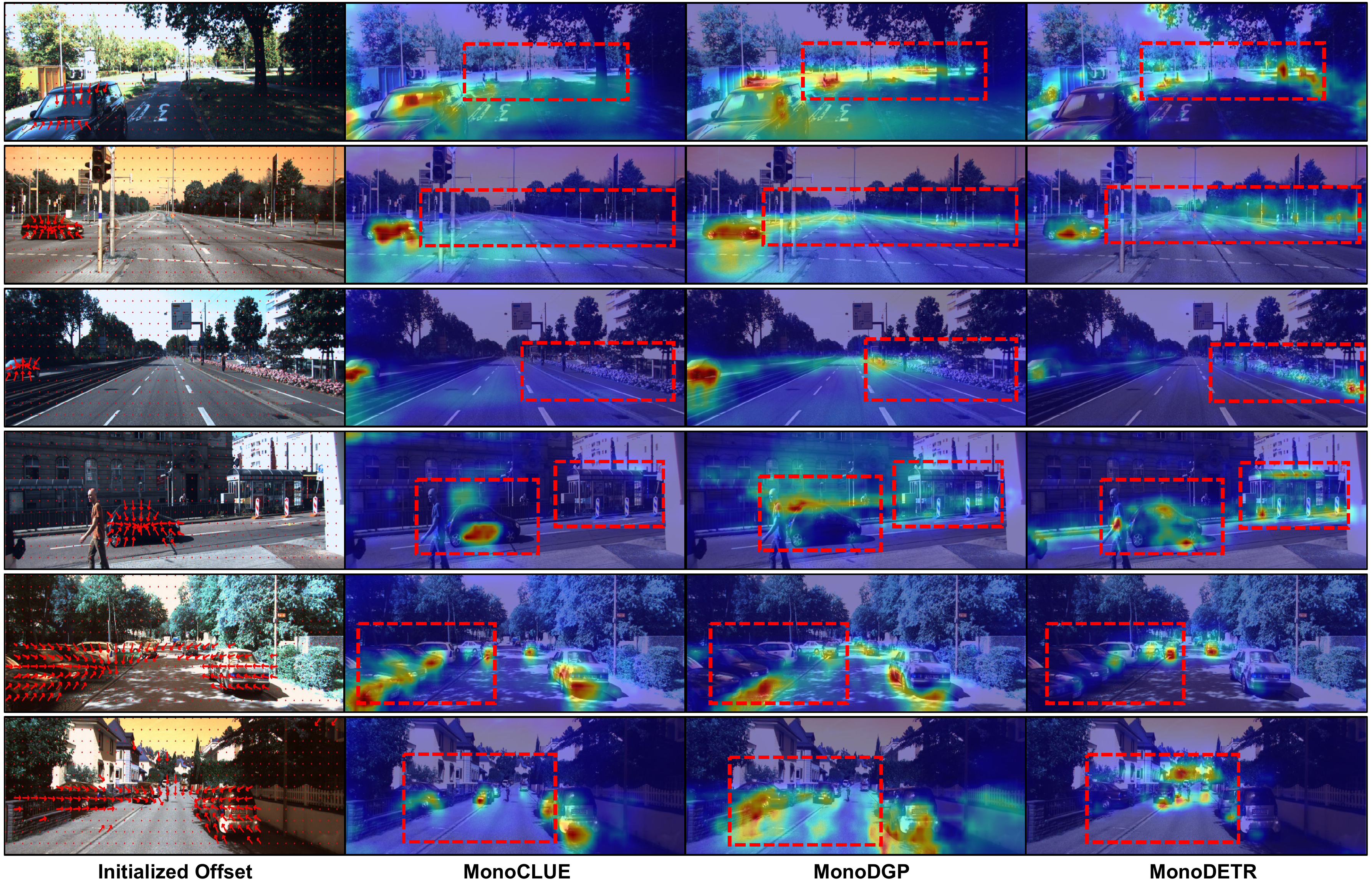}
	\captionof{figure}{Comparison of our model with offset initialization and previous models without it. The red box highlights a background region used to visualize the suppression of noise.}
	\label{fig:offset}
\end{strip}

\begin{figure*}[h]
	\centering
	\includegraphics[width=0.9\linewidth]{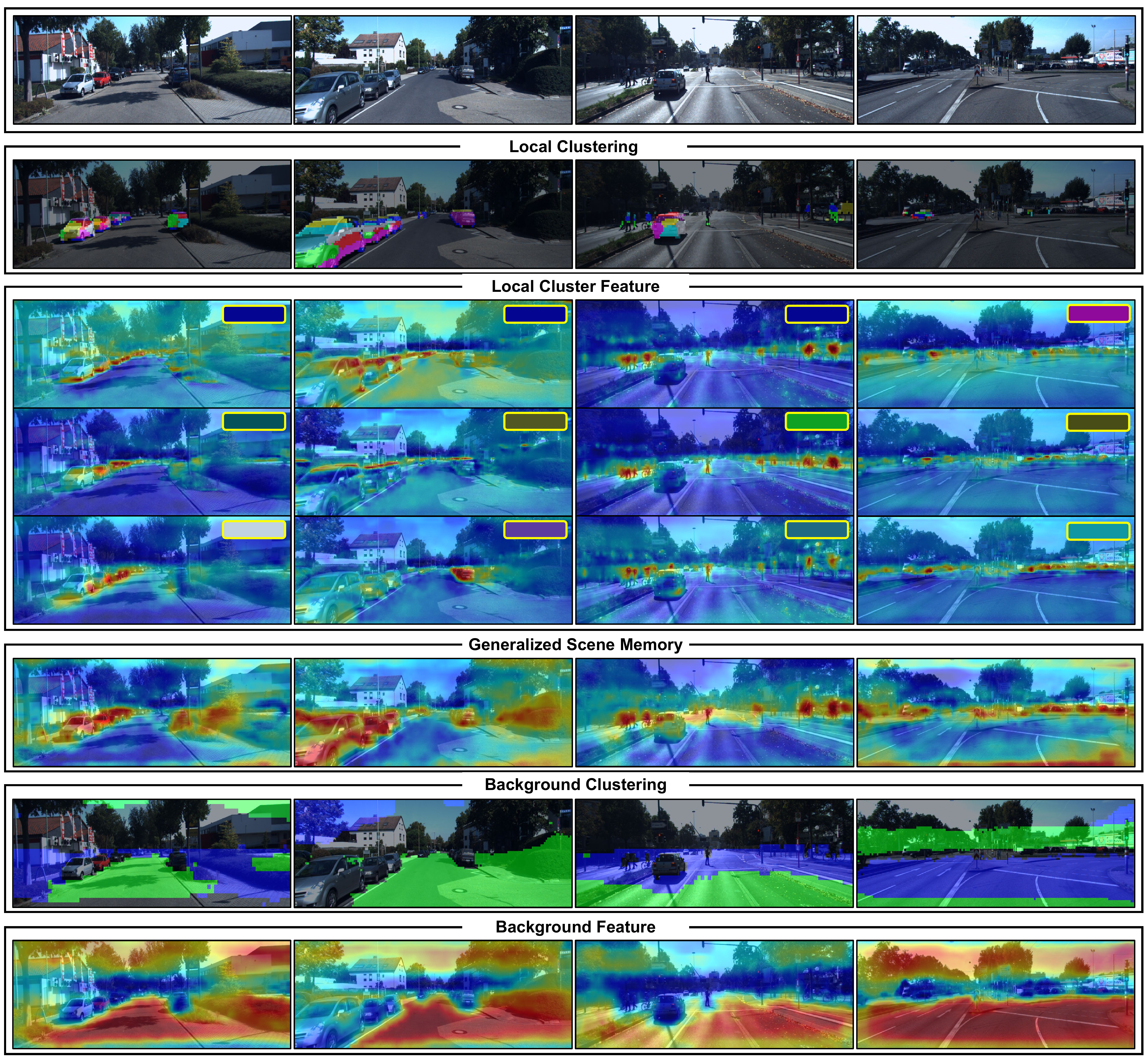}
	\caption{Visualization of overall clustering results.
Local clustering and background clustering show clustered regions, and each activation map represents the similarity to the full feature map $\textbf{F}_n^v$.}
	\label{fig:cluster}
\end{figure*}

\section{Details of Offset Initialization}
Previous methods~\cite{zhang2023monodetr, pu2025monodgp} utilize MS-Deformable Attention~\cite{zhu2020deformable} to flexibly attend to sparse and informative locations in multi-scale feature maps by sampling keypoints around each reference. To improve guidance toward object-centric regions, we initialize offsets using the similarity map $\textbf{S}$ as an auxiliary prior. This complements the learnable offsets and helps suppress background noise while improving attention stability.

First, we interpret the similarity map $\textbf{S}$ as a probability mass. For each feature level $n$, we apply a softmax function to produce a spatial weighting over $\textbf{S}_n(i,j)$. This process is expressed as follows:
\begin{equation}
\tilde{\textbf{S}}_n(i,j) = \frac{\exp(\textbf{S}_n(i,j))}{\sum_{k} \exp(\textbf{S}_n(i,j))},
\end{equation}
where $(i,j)$ indicate the pixel coordinates. 
Second, we estimate the center of attention by computing the weighted average of the grid positions $r_n(i,j) \in \mathbb{R}^{2 \times \frac{H}{n} \times \frac{W}{n}}$, where each $r_n(i,j)$ represents a 2D normalized coordinate at spatial location $(i,j)$. This process is expressed as follows:
\begin{equation}
c_n = \sum_{i,j} \tilde{\textbf{S}}_n(i,j) \cdot r_n(i,j).
\end{equation}
The $c_n \in \mathbb{R}^2$ represents an object-centric position computed as the weighted average of visual similarity scores, and serves to guide sampling points toward relevant object regions.
Third, we compute the offset between $c_n$ and each foreground reference point. This process is expressed as follows:
\begin{equation}
\Delta_n(i,j) = c_n - r_n(i,j)\cdot \tilde{\textbf{S}}_n(i,j).
\end{equation}
Here, the offset is modulated by the normalized similarity score to ensure that only confident matches contribute significantly.
Lastly, we initialize the reference points by adding the offset to the original reference point. This process is expressed as follows:
\begin{equation}
\tilde{r}_n(i,j) = r_n(i,j) + \Delta_n(i,j).
\end{equation}
These $\tilde{r}_n$ are then used as the reference points in ours MS-Deformable Attention.

This initialization brings several benefits. As the offset distribution is directed toward object-centric regions, the model attends less to background areas, leading to reduced noise from irrelevant regions. Furthermore, compared to random initialization, it enhances object-centric focus by clustering attention around the estimated object centers. By guiding the initial sampling locations toward likely object regions, the attention process becomes more structured. As shown in Figure~\ref{fig:offset}, the offset directions are oriented toward the centers of objects or object groups. Consequently, lower attention weights are observed in background regions within the MS-Deformable module, leading to more precise attention concentrated inside objects in MonoCLUE.

As a result, this initialization provides a reliable starting point for the learnable offsets, enabling the model to focus more effectively on informative regions.

\begin{table*}[!ht]
\small
\setlength{\tabcolsep}{4pt}
\centering
\renewcommand{\arraystretch}{1.0}
\setlength{\belowcaptionskip}{-5pt}
\begin{tabular}{l||ccc|ccc||ccc|ccc}
\toprule
\multirow{3}{*}{\textbf{Methods}} 
& \multicolumn{6}{c||}{\textbf{Test}} 
& \multicolumn{6}{c}{\textbf{Validation}} \\
\cmidrule{2-13}
& \multicolumn{3}{c|}{$AP_{BEV|R40}$} & \multicolumn{3}{c||}{$AP_{3D|R40}$}
& \multicolumn{3}{c|}{$AP_{BEV|R40}$} & \multicolumn{3}{c}{$AP_{3D|R40}$} \\
& Easy & Mod. & Hard & Easy & Mod. & Hard & Easy & Mod. & Hard & Easy & Mod. & Hard \\
\midrule\midrule

OPA-3D$^\dagger$~\cite{su2023opa} & 15.65 & 10.49 & 19.22 & 24.60 & 17.05 & 14.25 & 33.80 & 25.51 & 22.13 & 24.97 & 19.40 & 16.59 \\
\addlinespace[0.5pt]

MonoCon~\cite{liu2022learning} & 31.12 & 22.10 & 19.00 & 22.50 & 16.46 & 13.95 & - & - & - & 26.33 & 19.01 & 15.98 \\
DEVIANT~\cite{kumar2022deviant} & 29.65 & 20.44 & 17.43 & 21.88 & 14.46 & 11.89 & 32.60 & 23.04 & 19.99 & 24.63 & 16.54 & 14.52 \\
MonoDDE~\cite{li2022diversity} & 33.58 & 23.46 & 20.37 & 24.93 & 17.14 & 15.10 & 35.51 & 26.48 & 23.07 & 26.66 & 19.75 & 16.72 \\
\addlinespace[0.5pt]
MonoUNI~\cite{jinrang2023monouni} & 33.28 & 23.05 & 19.39 & 24.75 & 16.73 & 13.49 & - & - & - & 24.51 & 17.18 & 14.01 \\

MonoDGP~\cite{pu2025monodgp} & \underline{35.19} & \underline{24.50} & \underline{21.29} & \underline{26.05} & \underline{18.55} & \underline{15.95} & \underline{36.55} & \underline{27.51} & \underline{24.10} & \underline{29.41} & \underline{21.21} & \underline{18.11} \\
\addlinespace[0.5pt]
\midrule\midrule

MonoCLUE & \textbf{35.83} & \textbf{24.59} & \textbf{21.36} & \textbf{26.59} & \textbf{18.70} & \textbf{15.99} & \textbf{40.16} & \textbf{29.20} & \textbf{25.38} & \textbf{30.81} & \textbf{22.81} & \textbf{19.50} \\
\bottomrule
\end{tabular}

\caption{Comparisons with monocular methods evaluated on the KITTI validation and test sets after training on all categories. The best results are shown in \textbf{bold}, and the second-best are shown with \underline{underline}. $^\dagger$ denotes those using extra depth.}
\label{tab:tab1}
\end{table*}

\begin{table}[]
\small

\setlength{\tabcolsep}{4pt}
\centering
\renewcommand{\arraystretch}{1.0}
\setlength{\belowcaptionskip}{-5pt}
\begin{tabular}{l||ccc|ccc}
\toprule
\multirow{2}{*}{\textbf{Methods}} 
& \multicolumn{3}{c|}{\textbf{Pedestrian}} 
& \multicolumn{3}{c}{\textbf{Cyclist}} \\
& Easy & Mod. & Hard & Easy & Mod. & Hard \\
\midrule\midrule

GUPNet & 14.72 & 9.53 & 7.87 & 4.18 & 2.65 & 2.09 \\
MonoCon & 13.10 & 8.41 & 6.94 & 2.80 & 1.92 & 1.55 \\
DEVIANT & 13.43 & 8.65 & 7.69 & 5.05 & 3.13 & 2.59 \\
MonoDDE & 11.13 & 7.32 & 6.67 & \underline{5.94} & \underline{3.78} & \underline{3.33} \\
MonoUNI & \underline{15.78} & \underline{10.34} & \underline{8.74} & \textbf{7.34} & \textbf{4.28} & \textbf{3.78} \\
MonoDGP & 15.04 & 9.89 & 8.38 & 5.28 & 2.82 & 2.65 \\
\midrule\midrule
MonoCLUE & \textbf{16.18} & \textbf{10.45} & \textbf{8.75} & 
5.93 & 3.20 & 2.94 \\
\bottomrule
\end{tabular}
\caption{ Comparisons of other categories on
the KITTI test set.}
\label{tab:tab2}
\end{table}

\begin{table}[]
\centering
\small
\setlength{\belowcaptionskip}{-10pt}
\setlength{\tabcolsep}{2.5pt}
\begin{tabular}{c|ccc||c|ccc}
\toprule
\multirow{2}{*}{\#Clusters} & \multicolumn{3}{c||}{$\mathbf{L}_c$ / $AP_{3D|R40}$} & \multirow{2}{*}{\#Clusters} & \multicolumn{3}{c}{$\mathbf{B}_c$ / $AP_{3D|R40}$} \\
\cmidrule(lr){2-4} \cmidrule(lr){6-8}
& Easy & Mod. & Hard & & Easy & Mod. & Hard \\
\midrule
5  &   31.55   &   23.11   &  19.81  & 3  &   \textbf{33.74}   &   \textbf{24.10}  &   \textbf{20.58}      \\
10 &   \textbf{33.74}   &   \textbf{24.10}   &   \textbf{20.58}   & 5  &   32.04   &  23.44  & 19.77     \\
15 &    31.92  &    23.36 &    20.08 & 9  &  31.69    &   23.26   &    19.75  \\
\bottomrule
\end{tabular}
\caption{Comparison of cluster counts for local ($L_c$) and background ($B_c$) clustering on the KITTI val set.}
\label{tab:tab3}
\end{table}

\section{Additional Explanations of MonoCLUE}
The core contribution of our method lies in the use of clustering.
The two primary components are local cluster features $\textbf{L}_c$ and a generalized scene memory $\textbf{G}_c$, both of which enhance object-level representation.
Additionally, background cluster feature $\textbf{B}_c$ is incorporated to support the decoding process by providing contextual cues.

Figure~\ref{fig:cluster} presents a visualization of the entire set of features obtained through clustering. We visualize how $\textbf{L}_c$, $\textbf{G}_c$, and $\textbf{B}_c$ relate to the full visual feature map $\textbf{F}_n^v$ in terms of similarity. In local clustering, visual patterns with similar appearances are grouped into the same cluster, allowing the model to capture object-like regions that are not detected by the segmentation head.
Since this task involves 3D object detection, orientation is also considered, and objects facing opposite directions are assigned to different clusters.
Furthermore, $\textbf{L}_c$ is also applicable to pedestrian and cyclist categories.
The clustered features are separated into the head, upper body, and lower body, and each part shows high similarity across different instances of the person.
$\textbf{G}_c$ provides a common representation of overall object locations, capturing coarse object regions rather than fine-grained details.
Lastly, $\textbf{B}_c$ capture ground and surrounding contextual information. As shown in Figure 2, background features reveal that these features reflect the approximate structure of the ground.

As a result, our clustering method enables the use of diverse visual cues that previous methods cannot exploit.
This capability allows the model to maintain robust 3D object detection performance, even in challenging scenarios involving occlusion, truncation, or partially visible objects.

\begin{table*}[!ht]
\centering
\small
\setlength{\belowcaptionskip}{-10pt}
\renewcommand{\arraystretch}{1.0}
\begin{tabular}{l|c|cccc|cccc}
\toprule
\multicolumn{1}{l|}{\multirow{3}{*}{Methods}} & \multicolumn{1}{c|}{\multirow{3}{*}{Difficulty}} & \multicolumn{8}{c}{Validation} \\ \cmidrule{3-10}  
\multicolumn{1}{l|}{}  & \multicolumn{1}{c|}{} & \multicolumn{4}{c|}{$AP_{3D}$}  & \multicolumn{4}{c}{$APH_{3D}$} \\  
\multicolumn{1}{l|}{}   & \multicolumn{1}{c|}{}   & All & 0-30 & 30-50 & \multicolumn{1}{c|}{50-$\infty$} & All & 0-30 & 30-50 & 50-$\infty$ \\ \midrule
\multicolumn{1}{l|}{PCT$^*$~\cite{wang2021progressive}}   & \multicolumn{1}{c|}{\multirow{6}{*}{\makecell{Level~1\\ $(\text{IoU}=0.7)$}}}
 & 0.89 & 3.18 & 0.27 & \multicolumn{1}{c|}{0.07} & 0.88 & 3.15 & 0.27 & 0.07 \\
\multicolumn{1}{l|}{M3D-RPN~\cite{brazil2019m3d}}    & \multicolumn{1}{c|}{} & 0.35 & 1.12 & 0.18 & \multicolumn{1}{c|}{0.02} & 0.34 & 1.10 & 0.18 & 0.02 \\
\multicolumn{1}{l|}{GUPNet~\cite{lu2021geometry}}     & \multicolumn{1}{c|}{} & 2.28 & 6.15 & 0.81 & \multicolumn{1}{c|}{0.03} & 2.27 & 6.11 & 0.80 & 0.03 \\
\multicolumn{1}{l|}{DEVIANT~\cite{kumar2022deviant}}   & \multicolumn{1}{c|}{} & 2.69 & 6.95 & 0.99 & \multicolumn{1}{c|}{0.02} & 2.67 & 6.90 & \underline{0.98} & 0.02 \\
\multicolumn{1}{l|}{MonoUNI~\cite{jinrang2023monouni}}   & \multicolumn{1}{c|}{} & \underline{3.20} & \underline{8.61} & 0.87 & \multicolumn{1}{c|}{0.13} & \underline{3.16} & \underline{8.50} & 0.86 & 0.12 \\
\multicolumn{1}{l|}{MonoDGP~\cite{pu2025monodgp}}    & \multicolumn{1}{c|}{} & 2.41 & 6.67 & 0.84 &  \multicolumn{1}{c|}{0.12} & 2.39 & 6.62 & 0.84 &0.12 \\ 
\multicolumn{1}{l|}{MonoCoP}    & \multicolumn{1}{c|}{} & 2.72 & 7.44 & \textbf{1.07} & \multicolumn{1}{c|}{\textbf{0.16}} & 2.70 & 7.38 & \textbf{1.07} & \underline{0.16} \\ 
\midrule
\multicolumn{1}{l|}{MonoCLUE}   & \makecell{Level 1\\ $(\text{IoU}=0.7)$} & \textbf{3.32} & \textbf{8.87} & \underline{1.02} & \multicolumn{1}{c|}{\underline{0.15}} & \textbf{3.31} & \textbf{8.78} & \underline{0.98} & \textbf{0.18} \\ \midrule\midrule

\multicolumn{1}{l|}{PCT$^*$~\cite{wang2021progressive}}   & \multicolumn{1}{c|}{\multirow{6}{*}{\makecell{Level~2\\ $(\text{IoU}=0.7)$}}} & 0.66 & 3.18 & 0.27 & \multicolumn{1}{c|}{0.07} & 0.66 & 3.15 & 0.26 & 0.07 \\
\multicolumn{1}{l|}{M3D-RPN~\cite{brazil2019m3d}}    & \multicolumn{1}{c|}{} & 0.35 & 1.12 & 0.18 & \multicolumn{1}{c|}{0.02} & 0.33 & 1.10 & 0.17 & 0.02 \\
\multicolumn{1}{l|}{GUPNet~\cite{lu2021geometry}}     & \multicolumn{1}{c|}{} & 2.14 & 6.13 & 0.78 & \multicolumn{1}{c|}{0.02} & 2.12 & 6.08 & 0.77 & 0.02 \\
\multicolumn{1}{l|}{DEVIANT~\cite{kumar2022deviant}}   & \multicolumn{1}{c|}{} & 2.52 & 6.93 & 0.95 & \multicolumn{1}{c|}{0.02} & 2.50 & 6.87 & 0.94 & 0.02 \\
\multicolumn{1}{l|}{MonoUNI~\cite{jinrang2023monouni}}   & \multicolumn{1}{c|}{} & \underline{3.04} & \underline{8.59} & 0.85 & \multicolumn{1}{c|}{\underline{0.12}} & \underline{3.00} & \underline{8.48} & 0.84 & 0.12 \\
\multicolumn{1}{l|}{MonoDGP~\cite{pu2025monodgp}}    & \multicolumn{1}{c|}{} & 2.26 & 6.65 & 0.81 & \multicolumn{1}{c|}{0.10} & 2.24 & 6.59 & 0.81 & 0.10 \\ 
\multicolumn{1}{l|}{MonoCoP}    & \multicolumn{1}{c|}{} & 2.55 & 7.41 & \textbf{1.03} & \textbf{0.14} & 2.53 & 7.35 & \textbf{1.02} & \underline{0.14}  \\ 

\midrule
\multicolumn{1}{l|}{MonoCLUE}   & \makecell{Level 2\\ $(\text{IoU}=0.7)$} & \textbf{3.21} & \textbf{8.84} & \underline{0.98} & \multicolumn{1}{c|}{\textbf{0.14}} & \textbf{3.24} & \textbf{8.75} & \underline{0.95} & \textbf{0.15} \\ \bottomrule                                
\end{tabular}

\caption{Comparisons on the Waymo validation set for the car category. $^*$ indicates use of extra depth data.}
\label{tab:tab4}
\end{table*}

\section{Additional Experiments}

\subsection{Waymo}
We conduct experiments on the widely used Waymo Open Dataset~\cite{sun2020scalability}, a large-scale benchmark for 3D object detection in autonomous driving.
Following the DEVIANT split~\cite{kumar2022deviant}, we use 52,386 training images and 39,848 validation images sampled from the front camera.
Waymo defines two object levels (Level 1 and Level 2) based on the number of LiDAR points within each 3D bounding box.
Evaluation is performed across three distance ranges: $[0, 30), [30, 50)$ and $[50, \infty)$ meters, using both $AP_{3D}$ and $APH_{3D}$ metrics, the latter incorporating heading accuracy.

Table~\ref{tab:tab4} presents the performance of MonoCLUE on the Waymo validation set. Our model achieves state-of-the-art results in all distance ranges (All). In particular, MonoCLUE achieves the highest $AP_{3D}$ and $APH_{3D}$ scores in both Level 1 and Level 2 settings, with especially strong performance in the $[0, 30)$ range. Furthermore, MonoCLUE achieves robust performance in the long-range, such as $[50, \infty)$ meters. These improvements are achieved solely through enhanced visual representation, without relying on geometric enhancements.

\subsection{Other Categories on KITTI}
While most monocular 3D object detection methods primarily focus on the car category, some studies also consider other categories. Therefore, we include an ablation study comparing performance across different categories, as shown in Tables~\ref{tab:tab1} and~\ref{tab:tab2}. 

Table~\ref{tab:tab1} presents the car category results when trained jointly with other categories. Our method achieves state-of-the-art performance across all test set difficulty levels. Additionally, on the validation set, MonoCLUE shows notable improvements in $AP_{3D}$ with gains of +1.60\% for the moderate and +1.39\% for the hard cases. These results demonstrate that MonoCLUE maintains robust performance even when trained with additional categories. Table~\ref{tab:tab2} shows the performance on the other categories. For pedestrian, our method achieves state-of-the-art results across all difficulty levels, significantly outperforming the baseline MonoDGP~\cite{pu2025monodgp}. This indicates that our clustering strategy also benefits object categories other than car. In contrast, cyclist category performance remains limited. We anticipate this is because of the DETR-style decoder employed by both MonoDGP~\cite{pu2025monodgp} and our method. This architecture relies on attention-based decoding, which requires sufficient data. The scarcity of cyclist instances weakens attention weights, resulting in less reliable detection.

\subsection{Number of Each Component}
Table~\ref{tab:tab3} presents the results for different numbers of clusters used in each clustering method. 

In the case of local cluster features $\textbf{L}_c$, using a small number of clusters results in limited separation of visual cues, leading to suboptimal performance. Conversely, when the number of clusters is too large, it becomes difficult to form representative groupings, and the model may over-segment the scene into less meaningful regions, which also degrades performance. For background features $\textbf{B}_c$, increasing the number of clusters consistently reduces performance. This suggests that overly fine-grained partitioning of background regions may degrade the effectiveness of contextual understanding. It can also reduce the model's ability to focus on objects during query initialization, with background features acting more like noise. 

Based on extensive experiments, we empirically determined the appropriate number of clusters for both $\textbf{L}_c$ and $\textbf{B}_c$.

\subsection{Additional Qualitative results}
\begin{figure*}[t]
	\centering
    \vspace{-15pt}
	\includegraphics[width=\linewidth]{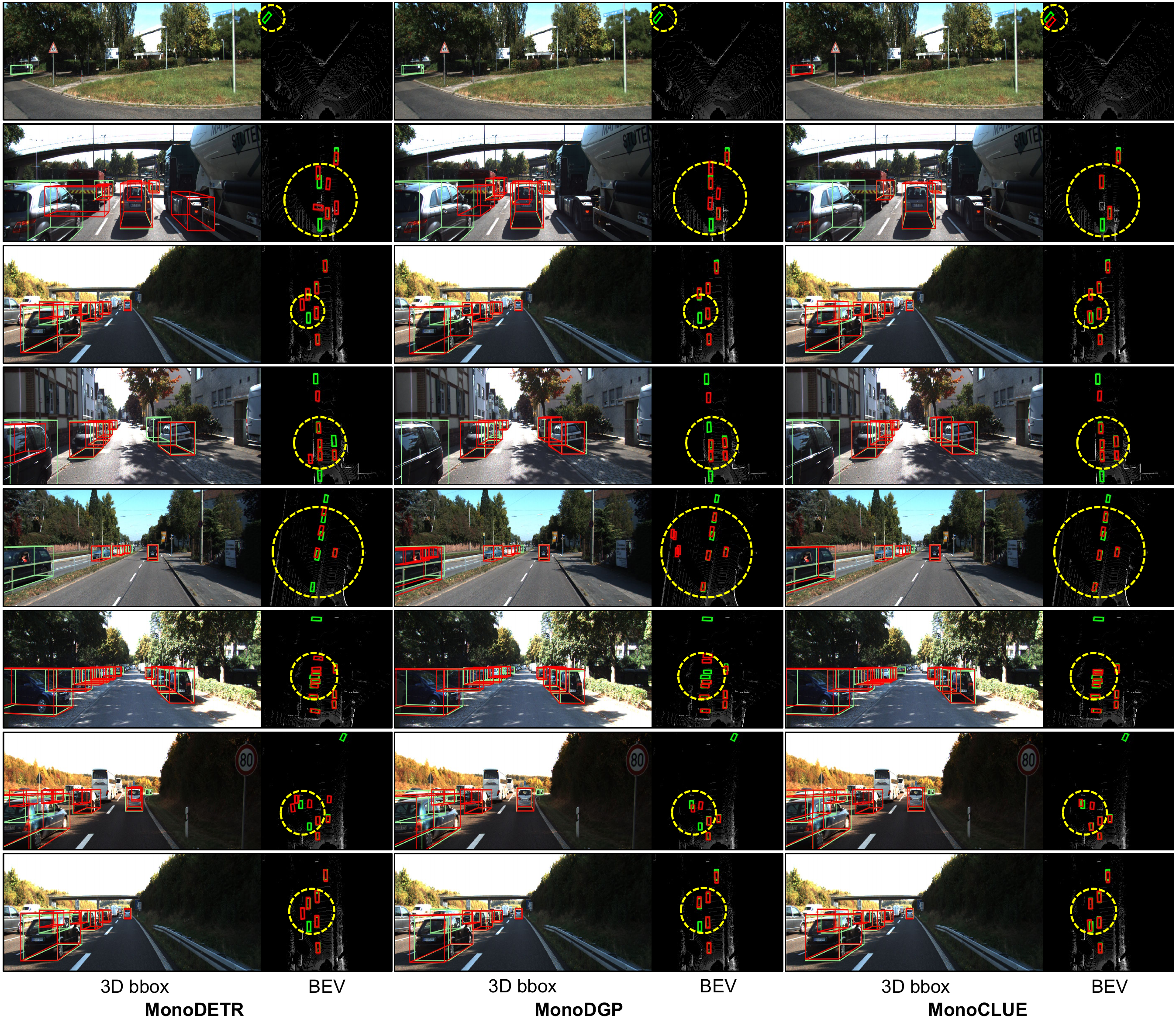}
	\caption{Additional qualitative results on the KITTI validation set. Red boxes indicate predictions, while green boxes indicate ground truth.}
	\label{fig:total}
\end{figure*}

Figure~\ref{fig:total} compares MonoCLUE with previous state-of-the-art methods under various scenarios.

MonoCLUE demonstrates superior performance compared to MonoDETR and MonoDGP, even in cases where objects are heavily occluded and only partially visible. This indicates that the similarity-based re-localization of local cluster features effectively handles partially visible regions. Moreover, MonoCLUE produces fewer outliers than the two baseline models. This is attributed to the query initializer, which incorporates local cluster features, generalized scene memory, and background features to embed object and scene-level priors. As a result, it enables the model to focus on relevant information based on high similarity, while suppressing responses to irrelevant regions.

\section{Limitations}
While MonoCLUE improves the utility of visual cues via clustering, its effectiveness depends on the availability of sufficient object instances. As discussed in Section 3.2, for categories with relatively few examples, such as cyclist, the benefits of clustering are less pronounced. In such cases, the small number of instances makes it difficult to capture consistent patterns in feature space, preventing the formation of reliable clusters and thereby limiting the benefit of clustering. This limitation also leads to weaker attention responses, resulting in less reliable detection. Furthermore, the model inherits certain limitations from the DETR architecture, which can lead to unstable training and high result variance, as also observed in other DETR-based monocular 3D object detection models. These issues hinder consistent and stable performance.

\bibliography{aaai2026}